\documentclass[letterpaper]{article} 
\usepackage{aaai24}  
\usepackage{times}  
\usepackage{helvet}  
\usepackage{courier}  
\usepackage[hyphens]{url}  
\usepackage{graphicx} 
\usepackage{amsmath}
\usepackage{amssymb}
\usepackage{booktabs}
\usepackage{multirow}
\usepackage{color}
\usepackage{natbib}  
\usepackage{caption} 
\usepackage{algorithm}
\usepackage{algorithmic}

\usepackage{newfloat}
\usepackage{listings}
\DeclareCaptionStyle{ruled}{labelfont=normalfont,labelsep=colon,strut=off} 
\floatstyle{ruled}
\newfloat{listing}{tb}{lst}{}
\floatname{listing}{Listing}

\title{MGNet:  Learning Correspondences via Multiple Graphs}
\author{
  Luanyuan Dai\textsuperscript{\rm 1},
 Xiaoyu Du\textsuperscript{\rm 1},
        Hanwang Zhang\textsuperscript{\rm 2}
 and Jinhui Tang\textsuperscript{\rm 1}\thanks{Corresponding author.}
}

\affiliations{
    \textsuperscript{\rm 1}Nanjing University of Science and Technology, China \\
    \textsuperscript{\rm 2}Nanyang Technological University, Singapore\\
   \{dailuanyuan, duxy, jinhuitang\}@njust.edu.cn, hanwangzhang@ntu.edu.sg\\
}

\usepackage{bibentry} 

\begin{document}

\maketitle

\begin{abstract}
Learning correspondences aims to find correct correspondences (inliers) from the initial correspondence set with an uneven correspondence distribution and a low inlier rate, which can be regarded as graph data. Recent advances usually use graph neural networks (GNNs) to build a single type of graph or simply stack local graphs into the global one to complete the task. But they ignore the complementary relationship between different types of graphs, which can effectively capture potential relationships among sparse correspondences. To address this problem, we propose MGNet to effectively combine multiple complementary graphs. To obtain information integrating implicit and explicit local graphs, we construct local graphs from implicit and explicit aspects and combine them effectively, which is used to build a global graph. Moreover, we propose Graph~Soft~Degree~Attention (GSDA) to make full use of all sparse correspondence information at once in the global graph, which can capture and amplify discriminative features.  Extensive experiments demonstrate that MGNet outperforms state-of-the-art methods in different visual tasks.  The code is provided in https://github.com/DAILUANYUAN/MGNet-2024AAAI.
\end{abstract}

\section{Introduction}

Finding high-quality pixel-wise correspondences is the precondition for many important computer vision and robotics tasks, $e.g.$, visual localization \cite{sattler2018benchmarking}, image stitching \cite{ma2019infrared}, image registration \cite{ma2015robust,9328884}, point cloud registration \cite{bai2021pointdsc,qin2022geometric}, Simultaneous Location and Mapping (SLAM) \cite{mur2015orb}, Structure from Motion (SfM) \cite{schonberger2016structure}, etc. A standard pipeline  depends on off-the-shelf detector-descriptors \cite{lowe2004distinctive,detone2018superpoint} to obtain putative correspondences, which have excessive incorrect correspondences  (i.e., outliers) due to the challenging cross-image variations, such as rotations, illumination changes and viewpoint changes.
\begin{figure} 
	\centering \label{gdt}
		\includegraphics[width=8.3cm]{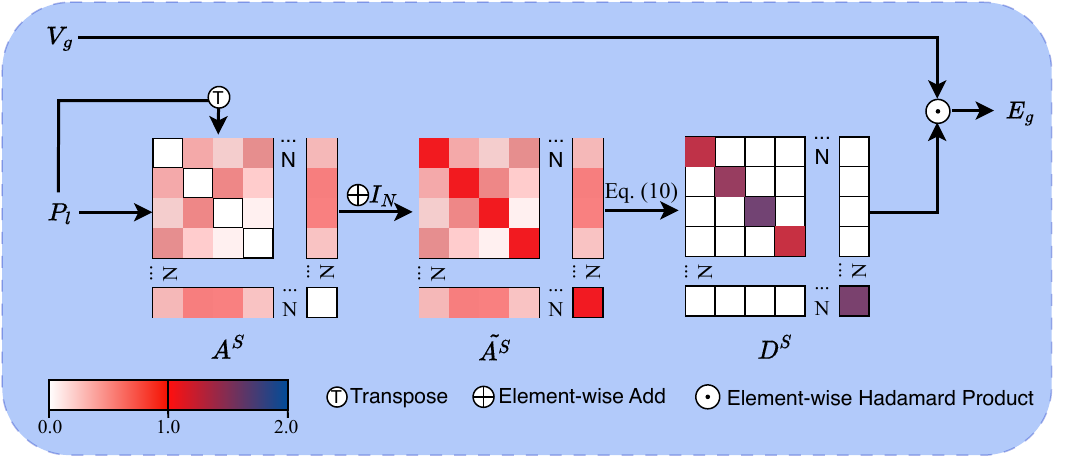}\\
	\caption{Graph~Soft~Degree~Attention, in which $A^S$, $\tilde{A^S}$ and $D^S$ represent Soft Adjacent Matrix, the final Soft Adjacent Matrix and Soft Degree Matrix, respectively. Combing with the like-probability value (white to red and then to blue is from 0 to 1 and then to 2), it can prove that Soft Degree Matrix	 $D^S$ can capture and amplify discriminative features.	 }\label{gdt}
\end{figure}

Hence, outlier rejection is an essential step to preserve correct correspondences as well as reject false ones. Initial correspondences are spread unevenly over an image pair, due to densely detected keypoints in textured areas but almost no keypoints in textureless areas. Hence, some networks \cite{zhang2019learning,liu2021learnable,zhao2021progressive,Dai_2022_CVPR,citation-0} view sparse correspondences as  graph data, in which there is no order and unified structure. OA-Net \cite{zhang2019learning}, U-Match \cite{citation-0} and MS$^{2}$DG-Net \cite{Dai_2022_CVPR} only construct graphs in the local region without considering the global, where the first two implicitly construct local graphs and the other adopts an explicit approach. CL-Net \cite{zhao2021progressive} simply stacks explicit local graphs into the global one, which is coped with a plain spectral graph convolutional layer (GCN) \cite{kipf2016semi}. At the same time, LMC-Net \cite{liu2021learnable} only builds global graph Laplacian based on standard Laplacian matrix and decomposes it to solve the proposed formulation. They fail to consider potential relationships among different types of graphs and how to effectively use Laplacian matrix on graph data.

These networks have made certain progress in handling sparse correspondences, but there are still some problems. Firstly, no one uses GNNs to construct graphs from implicit and explicit perspectives at the same time, and explore their relationships and complementary advantages. Secondly, the ability of the plain spectral graph convolutional layer (GCN) \cite{zhao2021progressive,kipf2016semi}  is not strong enough to capture discriminative feature in the global graph. That is to say, mainstream methods do not make full use of GNNs on sparse correspondences. Therefore, we propose a network, named MGNet, which effectively combines multiple graphs, to handle these sparse correspondences. Firstly, we build local graphs through implicit and explicit perspectives at the same time by GNNs, and explore potential relationships between them. Then, we propose Graph~Soft~Degree~Attention (GSDA) to obtain and amplify discriminative features in the global graph.  As shown in Figure \ref{gdt}, Soft Adjacent Matrix $A^S$ does not consider its own information, and the final Soft Adjacent Matrix $\tilde{A^S}$ pays little attention to relationships between the selected sparse correspondence and others. In Soft Degree Matrix $D^S$, inspired by Laplace matrix, each selected correspondence fuses relationships between itself and all other correspondences. Hence,  GSDA can capture and amplify discriminative features, as shown in Figure \ref{gdt}.  

Our contribution is threefold. Firstly, implicit and explicit graphs are constructed at the same time by GNNs, and  potential relationships between them have been discussed at length. After that, motivated by Laplacian matrix, Graph~Soft~Degree~Attention (GSDA) is proposed and applied  to effectively handle global information at once in the global graph, which can capture and amplify discriminative features.  Finally, the proposed MGNet  obtains state-of-the-art results on 
camera pose estimation, homography estimation, and visual localization with a relatively small number of parameters.

\section{Related Work}
\subsection{Outlier~Rejection} \label{tor}
Traditional RANSAC \cite{fischler1981random} and its variants \cite{torr1998robust,chum2005two,barath2018graph,barath2019magsac,barath2020magsac++}  capture correct correspondences via the largest subset, so they  may conform to specific scenarios. Thus, with the increasing of general dataset scale and outlier ratio, nearly all of them no longer work. Hence, using deep learning-based networks to handle irregular and unordered characteristics among sparse correspondences has emerged. First, CNe \cite{moo2018learning} and DFE \cite{ranftl2018deep} only take correspondence coordinates as input and achieve great success. After that, some networks introduce the thought of  attention mechanism \cite{vaswani2017attention} to enhance network performance. ACNe \cite{sun2020acne} and LAGA-Net \cite{dai2021enhancing} exploit attention mechanisms from local and global perspectives, but use different approaches. ANA-Net \cite{ye2023learning} provides the idea of second-order attention and proves its existence. To search more reliable correspondences, LFLN-Net \cite{Wang2020} and NM-Net \cite{zhao2019nm} redefine neighborhood from different aspects. Next,  LMC-Net \cite{liu2021learnable} utilizes consistency constraint to remove outliers.  CL-Net \cite{zhao2021progressive} introduces a pruning operation to obtain inlier identification.

\subsection{Graph~Neural~Network~in~Correspondences} \label{gnn}

Recently, Graph~Neural~Networks (GNNs) have emerged in correspondence learning, due to their powerful feature extraction abilities.   To our knowledge, OA-Net \cite{zhang2019learning}  is the first one introducing GNNs to remove outliers in sparse correspondences, which is inspired by DIFFPOOL \cite{ying2018hierarchical} and improve DiffUnpool from plain to order-aware by a soft assignment manner. In LMC-Net \cite{liu2021learnable}, Liu et al. introduce graph Laplacian to decompose a new proposed formulation of motion coherence among sparse correspondences. In CL-Net \cite{zhao2021progressive}, Zhao et al. rely on dynamic graphs to obtain local and global consensus scores to progressively choose correct correspondences, where an annular convolutional operation is proposed to fuse local features. In MS$^{2}$DG-Net \cite{Dai_2022_CVPR}, Dai et al. combine dynamic graphs and attention mechanism to capture local topology through similar sparse semantics information  in each image pair.  U-Match \cite{citation-0} combine a U-shaped network and GNNs, which can better utilize hierarchical graph information, to increase network ability to capture features. 

\section{Proposed Method}
\subsection{Problem~Formulation} \label{pr}
\begin{figure}
	\centering
	\includegraphics[width=8cm]{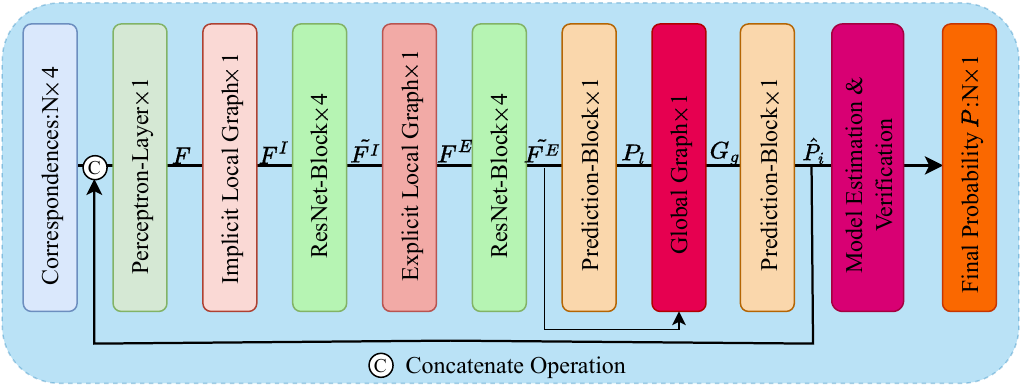}\\
	\caption{Network architecture of MGNet. The input is a putative correspondence set $C$, and the output is the final probability set  $P$. $i$ = $1$, $2$. }\label{arch}
\end{figure}
We use  local features (SIFT \cite{lowe2004distinctive},  SuperPoint \cite{detone2018superpoint}, etc.) followed by a NN matcher to build an initial correspondence set $C$. 
\begin{equation}
	C=\left\lbrace c_1;c_2;...;c_N\right \rbrace \in\mathbb{R}^{N\times4}
\end{equation}  
where $c_i=\left(x_i,y_i,u_i,v_i\right)$  is  a correspondence between two keypoints $\left(x_i,y_i\right)$ and $\left(u_i,v_i\right)$, which are normalized under camera intrinsics. The  initial correspondence set is polluted by excessive outliers, which can bring negative impact on downstream tasks.

Hence, we propose MGNet to reject outliers. Motivated by \cite{fischler1981random,zhao2021progressive}, we use a verification framework, but without the pruning operation \cite{zhao2021progressive}. That is because the pruning operation may reduce data abundance, and we prove it in Table \ref{pruning}. As shown in Figure \ref{arch}, we iteratively use our main network twice to obtain the final probability set  $P=\left\lbrace p_1;p_2;...;p_N\right\rbrace$ with $p_i\in [0,1)$, in which elements present probabilities of the whole correspondences as inliers. From the first iteration, we can obtain the first estimated inlier probability set $\hat{P}_{1}$. After that, we use $\hat{P}_{1}$ and $C$ to obtain $\hat{P}_{2}$ through the second iteration. Next, we use a weighted eight-point algorithm \cite{moo2018learning} to estimate an essential matrix $\hat{E}$. Finally, a  verification operation is used to test and verify  the estimated essential matrix $\hat{E}$ on the correspondence set $C$ and obtain the final probability set  $P$. 
\begin{equation}
 \begin{aligned}  \hat{P}_{1} & =f_{1\phi}(C), \quad \hat{P}_{2}=f_{2\psi}(\hat{P}_1,C) \\ \hat{E} & =g(\hat{P}_{2}, C), \quad P=Ver(\hat{E}, E),\end{aligned} 
\end{equation}
where $f_{1\phi}(\cdot)$ and $f_{2\psi}(\cdot,\cdot)$ represent the first and second iterations with learnable parameter $\phi$ and $\psi$, respectively; $\hat{P}_{1} $ and  $\hat{P}_{2}$ are the estimated inlier probability sets in the first and second iterations, respectively; $g(\cdot,\cdot)$ is the weighted eight-point algorithm; $Ver(\cdot,\cdot)$ is the verification operation.
\subsection{Implicit~and~Explicit~Local~Graphs}
\subsubsection{Build~Implicit~Local~Graph.}
Firstly, the input correspondence set $C$ is encoded into a $S$-dimensional feature set $F=\left\lbrace f_i\right\rbrace^N_{i=1}\in\mathbb{R}^{S\times N \times 1}$ by a Perceptron Layer. After that, DiffPooling operation \cite{zhang2019learning} is used to coarsen $F$ into a $M$-dimensional coarse-grained graph set $G^I=\left\lbrace g^I_i\right\rbrace^M_{i=1}\in\mathbb {R}^{S\times M \times 1}$ via an implicit way. Following OANet \cite{zhang2019learning}, we choose OA Filtering operation to process coarse-grained graphs, so that global information among them can be attained. Finally, DiffUnpooling operation \cite{zhang2019learning} is used to restore data to its original size by a soft way. These  can be written as: 
\begin{equation}	
	G^I = DiffPooling\left(F\right)
\end{equation}
\begin{equation}	
	F^I =DiffUnpooling\left(F, OA\left(G^I \right)\right)
\end{equation}
where  $DiffPooling(\cdot)$, $DiffUnpooling(\cdot,\cdot)$ and $OA(\cdot)$ represent DiffPooling, DiffUnpooling and OA Filtering operations, respectively; $F^I$ is denoted as an implicit local graph feature set.
\subsubsection{Build~Explicit~Local~Graph.}
First, we use ResNet blocks to extract an implicit local graph feature vector $\tilde{F^I}$ from $F^I$ and use it to construct explicit local graphs. Second, $k$-nearest neighbors are chosen in $\tilde{F^I}$ in feature space. After that, an edge set $E^E_{j}$ is constructed by concatenating the selected correspondence feature map and the residual ones in $\tilde{F^I}$, just like \cite{Dai_2022_CVPR,zhao2021progressive}. Next, an explicit graph set $G^E=\left\lbrace g^E_i\right\rbrace^N_{i=1}\in\mathbb {R}^{S \times N\times k}$ is built on $\tilde{F^I}$ with its $k$-nearest neighbors, to capture local topology among sparse correspondences. Finally, we choose maximum pooling and MLPs to aggregate information to obtain an explicit local graph feature set  $F^E\in\mathbb {R}^{S \times N\times 1}$. The above operations can be recorded as:
\begin{equation}	
	E^E = [\tilde{F^I_j}||\tilde{F^I} - \tilde{F^I_j}], j = {1,2,...,k}
\end{equation}
\begin{equation}	
	G^E =\left( V^E , E^E \right)
\end{equation}
\begin{equation}	
	F^E=maxpooling  \left( MLPs \left( G^E \right) \right)\end{equation}
where $[\cdot||\cdot]$ presents concatenation; $\tilde{F^I}$, $\tilde{F^I}_{j}$ and $\tilde{F^I} - \tilde{F^I}_{j}$ are correspondence, neighborhood and residual  feature sets, respectively;  $V^E$ indicates a $\tilde{F^I}$'s neighbor set; $ E^E$ denotes an explicit edge set.
\subsubsection{Relationship~between~them.}
The coarsening process of implicit local graphs is automatically learned and sparse correspondences can be automatically grouped. That is, the information of nodes (sparse correspondences) can be learned, and the local structural information (relationship among sparse correspondences) can also be learned at the same time. In addition, constructing explicit local graphs on the implicit local graph feature vector $\tilde{F^I}$, allows us to intuitively obtain the more accurate local explicit graphs, as shown in Table \ref{relationship}. Comparing with Table \ref{relationship} and the third, fourth and fifth lines in Table \ref{final}, we find that building implicit graphs first performs best. This may because mining the information among sparse correspondences from an implicit perspective first and then using the captured information to construct graphs from an explicit aspect can more fully explore the potential information and relationships among sparse correspondences.

\subsection{Construct~Global~Graph}
First, $F^E$ is put into ResNet blocks and an explicit local graph feature vector $\tilde{F^E}$ is obtained. We put $\tilde{F^E}$ into a Prediction layer so that we can obtain a local probability set $P_l$. Next, $\tilde{F^E}$ can be denoted as   the global graph node set $V^g$. After that, we propose a novelty yet simple (without additional parameters) approach,  named Graph~Soft~Degree~Attention (GSDA) to construct the global edge set $E^g$, as shown in Figure \ref{gdt}. Specifically, we explore relationships in every two members in the local probability set $P_l$ to produce Soft Adjacent Matrix  $A^S\in \mathbb {R}^{N \times N }$(see Theorem 1.), which cannot consider its own information, so a self-loop is created on top of it. The above operations can be written as:
\begin{equation}	
	A^S = softmax \left( P_l \cdot P^T_l\right)
\end{equation}
\begin{equation}	
	\tilde{A^S} = A^S + I_N
\end{equation}
where $I_N$ is a $N \times N$ unit matrix; $\tilde{A^S} =\left \lbrace \tilde{A^S}_{i,j} \right\rbrace^N_{i,j=1} \in \mathbb {R}^{N \times N}$ is the final Soft Adjacent Matrix.
 
After that, we construct Soft Degree Matrix $D^S =\left\lbrace D^S_{i,j} \right\rbrace^N_{i,j=1} \in \mathbb {R}^{N \times N }$, in which one diagonal element is the sum of the corresponding rows on the final Soft Adjacent Matrix $\tilde{A^S}$, and the remains are zeros. (See Theorem 2$\&$3.) One element on the Soft Degree Matrix $D^S$ diagonal represents the sum of relationships between the selected correspondence and others in an image pair, which can make full use of all sparse correspondence information at once and at a long distance. Comparing with $A^S$, $\tilde{A^S}$ and  $D^S$ visualizations in Figure \ref{gdt}, we find the proposed GSDA can capture and amplify discriminative features. That can be written as: 
 \begin{equation}
	D^S_{i,j}=\left\{
	\begin{aligned}
		\sum_{j=1}^N \tilde{A^S}_{i,j} & , & i=j \\
		0 & , & else
	\end{aligned}
	\right.
\end{equation}

Next, an element-wise Hadamard product is performed between Soft Degree Matrix $D^S$ and global graph node set $V^g$. Finally, a global graph is built by integrating implicit and explicit local graph information. These are defined as:
\begin{equation}	
	E^g = D^S \odot  V^g
\end{equation}\begin{equation}	
	G^g = \left( V^g, E^g \right)
\end{equation}
where $\odot$ is the element-wise Hadamard product.

Similar to the local probability set, the global  probability set $P_g$ ($\hat{P}_{i}$, $i$ = $1$, $2$) is defined by encoding the aggregated features by a ResNet block and a Prediction layer.

\subsection{Related~Theorem}
\subsubsection{Theorem~1.}
Adjacency Matrix $A \in \mathbb {R}^{N \times N }$ represents connections between any two nodes in graph data. If there is a connection between nodes $v_i$ and $v_j$, an edge $\left(v_i, v_j\right)$ will form and the corresponding element of Adjacency Matrix $A_{ij}=1$, otherwise $A_{ij}=0$. In addition, the diagonal element of Adjacency Matrix $A$ is usually set to $0$.
\subsubsection{Theorem~2.}
Degree of a node refers to the total number of edges connected to it.  $d \left( v \right)$ is usually used to present degree of a node.
\subsubsection{Theorem~3.}
Degree Matrix  $D =\left\lbrace d_{i,j} \right\rbrace^N_{i,j=1}$ of graph $G$ is an $N \times N$ diagonal matrix, and an element on the diagonal is degree of the corresponding node, represented as:
 \begin{equation}
	d_{i,j}=\left\{
	\begin{aligned}
		d \left( v_i \right)& , & i=j \\
		0 & , & else
	\end{aligned}
	\right.
\end{equation}
\subsection{Loss~Function}
Following OA-Net++ \cite{zhang2019learning} and CL-Net \cite{zhao2021progressive}, we choose a hybrid loss function:
\begin{equation} \label{loss}
	L = L_c+ \beta L_e(E,\hat{E})
\end{equation}
where $L_c$ is defined as a binary classification loss with a proposed adaptive temperature, provided by CL-Net \cite{zhao2021progressive};  the later is a geometric loss between  the ground truth $E$ and an predicted estimated model $\hat{E}$;  $\beta$ is a weighting  factor to balance both of them.

\subsection{Implementation Details} \label{id}
Network input is $N \times 4$ initial correspondences by SIFT or SuperPoint, and typically $N$ is up to $2000$. Cluster number $m$, neighbor number $k$ and channel dimension $S$ are $100$,  $24$ and $128$. Batchsize and $\beta$ in Equation \ref{loss} are set to $32$ and $0.5$, respectively. Adam \cite{paszke2017automatic} optimizer is used with a learning rate of $10^{-3}$ and we choose a warmup strategy. Clearly,  a linearly growing rate is used  for the first $10k$ iterations, after that  the learning rate begins to decrease and reduce for every $20k$ iterations with a factor of $0.4$. Experiments are performed on NVIDIA GTX 3090 GPUs.

\section{Experiments}
\subsection{Evaluation~Protocols} \label{ep}
\subsubsection{Main~Datasets.}
Yahoo's YFCC100M \cite{thomee2016yfcc100m} and SUN3D \cite{xiao2013sun3d} datasets are chosen as outdoor and indoor scenes, respectively. Following OA-Net++ \cite{zhang2019learning}, $68$ sequences are selected as training sequences and the rest $4$ sequences are regarded as unknown scenes in outdoor scenes, and $239$ sequences are chosen as training sequences, and the rest $15$ sequences are unknown scenes in indoor scenes. Incidentally, we divide training sequences into three parts, consisting of training ($60\%$), validation($20\%$) and testing ($20\%$), and the last one is used as known scenes.
\subsubsection{Main~Evaluation~Metrics.}
The error metrics can be defined by angular differences between calculated rotation/translation vectors (recovered from the essential matrix) and the ground truth. mAP$5^{\circ}$ and  mAP$20^{\circ}$ are selected  as the default metrics in the camera pose estimation task.

\subsection{Main~Baselines}
We choose a traditional  method (RANSAC \cite{fischler1981random}) and ten learning-based networks ( Point-Net++ \cite{qi2017pointnet++}, DFE \cite{ranftl2018deep}, CNe \cite{moo2018learning}, OA-Net++ \cite{zhang2019learning}, ACNe \cite{sun2020acne}, SuperGlue \cite{Sarlin_2020_CVPR}, LMC-Net \cite{liu2021learnable},  CL-Net \cite{zhao2021progressive}, MS$^{2}$DG-Net \cite{Dai_2022_CVPR} and U-Match \cite{citation-0}) as main baselines. The official SuperGlue model is directly used to test. 

\subsection{Camera~Pose~Estimation}
Camera pose estimation, referring to utilize the identified inliers to accurately excavate the relative position relationship (rotation and translation) between different camera views, is an important foundation for many computer vision tasks.
\begin{table}[]
	\centering
	\footnotesize
	\begin{tabular}
		{ccccc}
		\hline
		\multirow{2}*{Method}  &\multicolumn{2}{c}{Outdoor(\%)} &\multicolumn{2}{c}{Indoor(\%)}  \\
		\cline{2-5}
		&\multicolumn{1}{c}{Known}&\multicolumn{1}{c}{Unknown}&\multicolumn{1}{c}{Known}&\multicolumn{1}{c}{Unknown}
		\\
		\cline{2-5}
		\hline
		RANSAC&   5.81   &  9.07&   4.52   &  2.84 \\	
		Point-Net++ &  10.49  &  16.48&10.58& 8.10\\
		DFE & 19.13 &  30.27& 14.05& 12.06\\
		CNe &  13.81&   23.95&  11.55&  9.30 \\			
		OA-Net++&   32.57 & 38.95&   20.86&16.18\\
		ACNe& 29.17  & 33.06&18.86& 14.12\\
		SuperGlue &35.00&48.12&22.50&17.11\\
		LMC-Net &33.73&47.50&19.92&16.82\\
		  CL-Net&39.16&53.10&20.35&17.03\\  
		  MS$^{2}$DG-Net & 38.36  &  49.13&22.20  &   17.84     \\
		  U-Match&46.78&60.53&24.98&21.38	\\	
		  \cline{1-5}
		MGNet &  \textbf{51.43} &  \textbf{64.63} & \textbf{25.96}  &  \textbf{21.27}      \\					
		\hline	
	\end{tabular}
	\centering
	\caption{Evaluation on outdoor and indoor datasets with SIFT for camera pose estimation. The mAP$5^{\circ}(\%)$ is reported and best result in each column is bold.  }
	\label{sift}
\end{table}

\subsubsection{Camera~Pose~Estimation~Results~with~SIFT.} 
\begin{figure}[t]
	\centering
	\begin{minipage}[t]{.1\textwidth}
		\centering
		\centerline{\includegraphics[width=2cm, height=3.65cm]{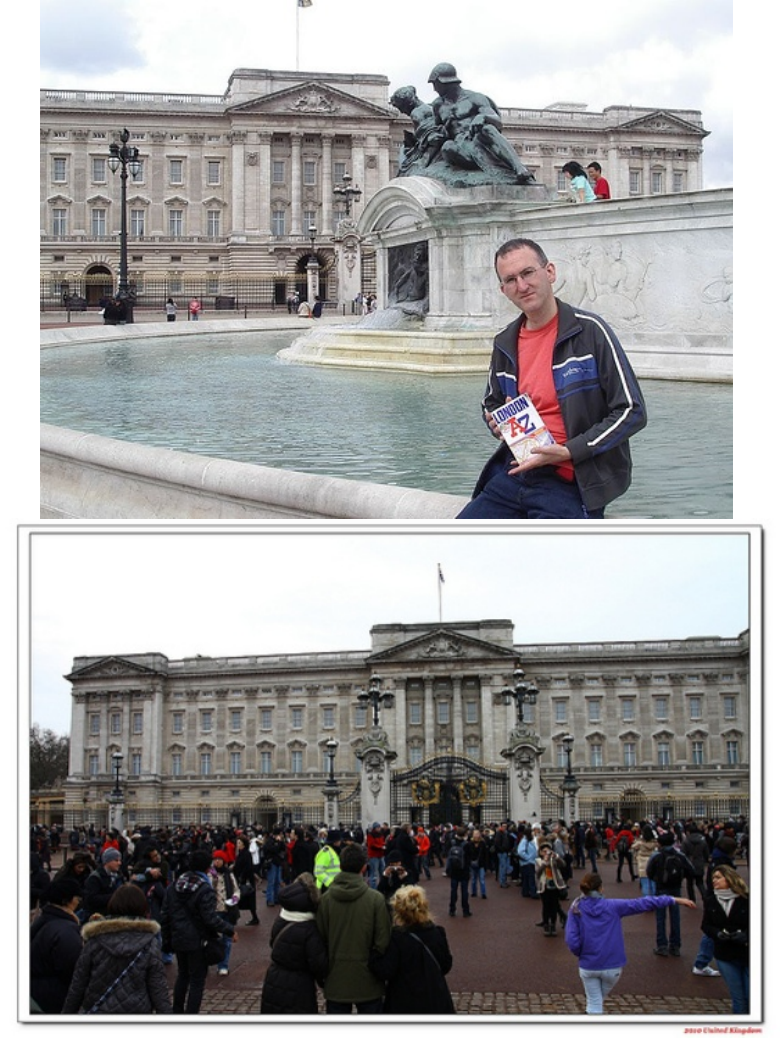}}
		\centerline{\includegraphics[width=2cm, height=3.65cm]{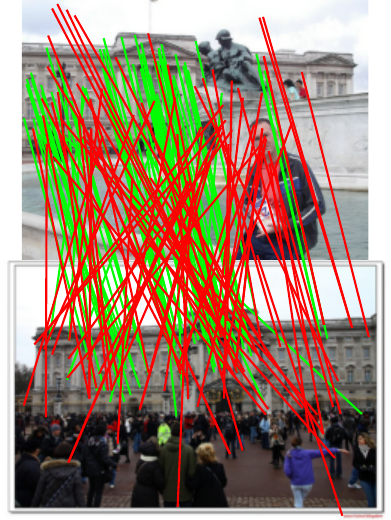}}
		\centerline{\includegraphics[width=2cm, height=3.65cm]{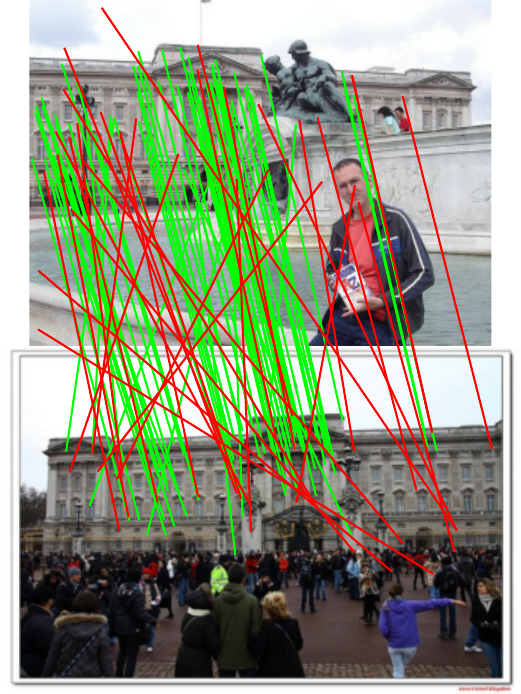}}
	\end{minipage}
	\begin{minipage}[t]{.1\textwidth}
		\centering
		\centerline{\includegraphics[width=2cm, height=3.65cm]{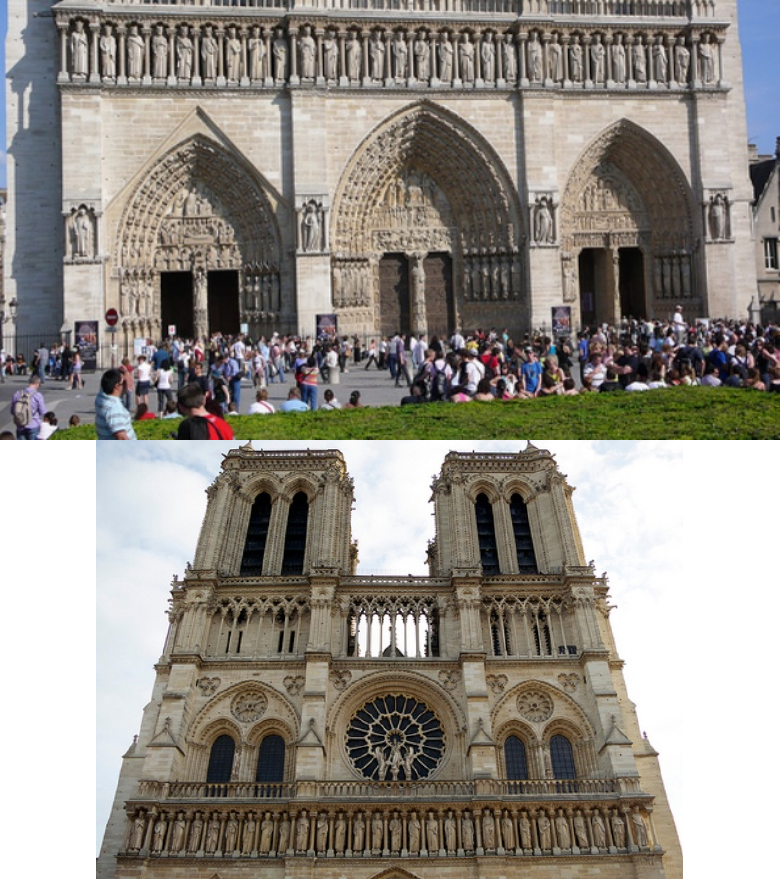}}
		\centerline{\includegraphics[width=2cm, height=3.65cm]{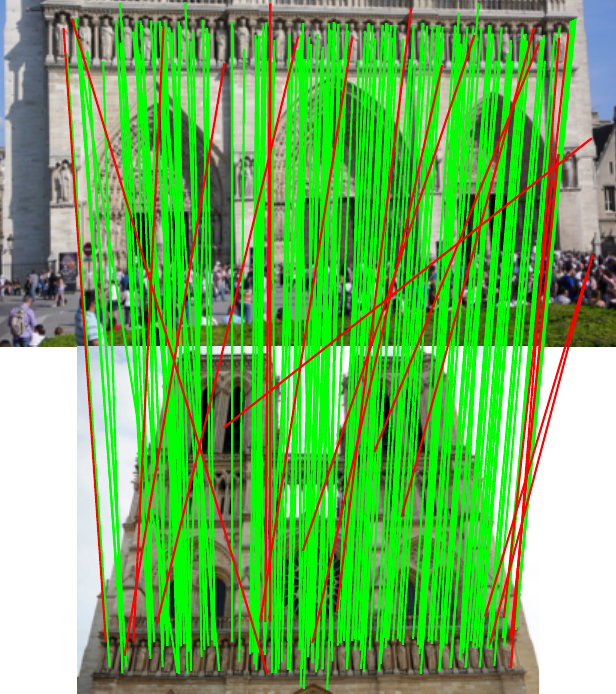}}
		\centerline{\includegraphics[width=2cm, height=3.65cm]{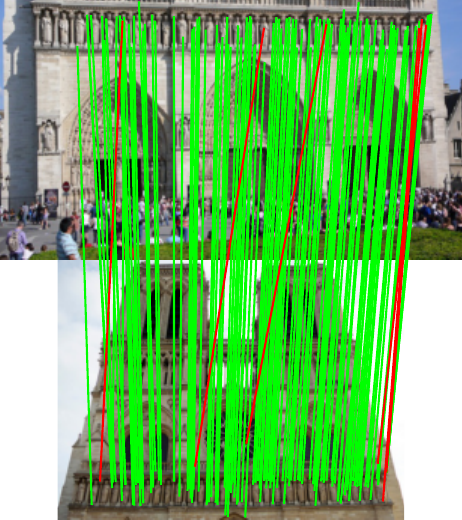}}
	\end{minipage}
	\begin{minipage}[t]{.1\textwidth}
		\centering
		\centerline{\includegraphics[width=2cm, height=3.65cm]{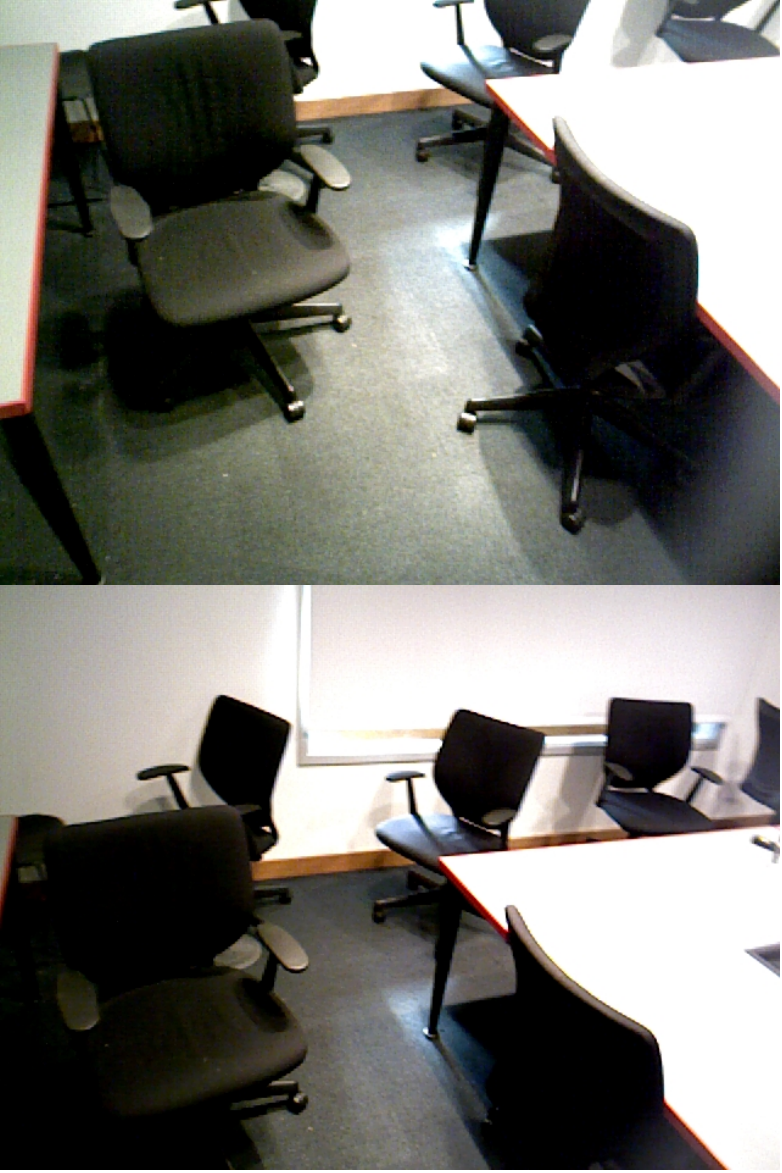}}
		\centerline{\includegraphics[width=2cm, height=3.65cm]{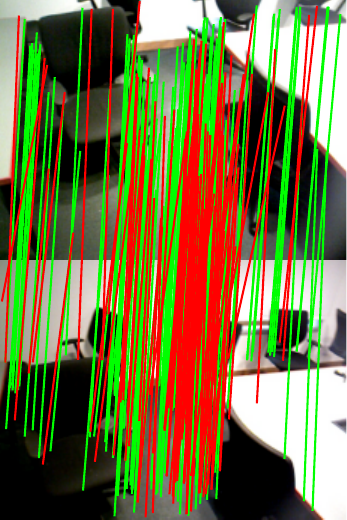}}
		\centerline{\includegraphics[width=2cm, height=3.65cm]{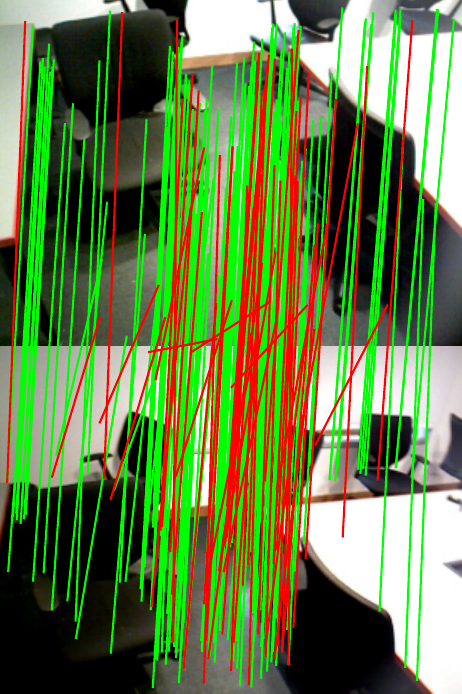}}
	\end{minipage}
	\begin{minipage}[t]{.1\textwidth}
		\centering
		\centerline{\includegraphics[width=2cm, height=3.65cm]{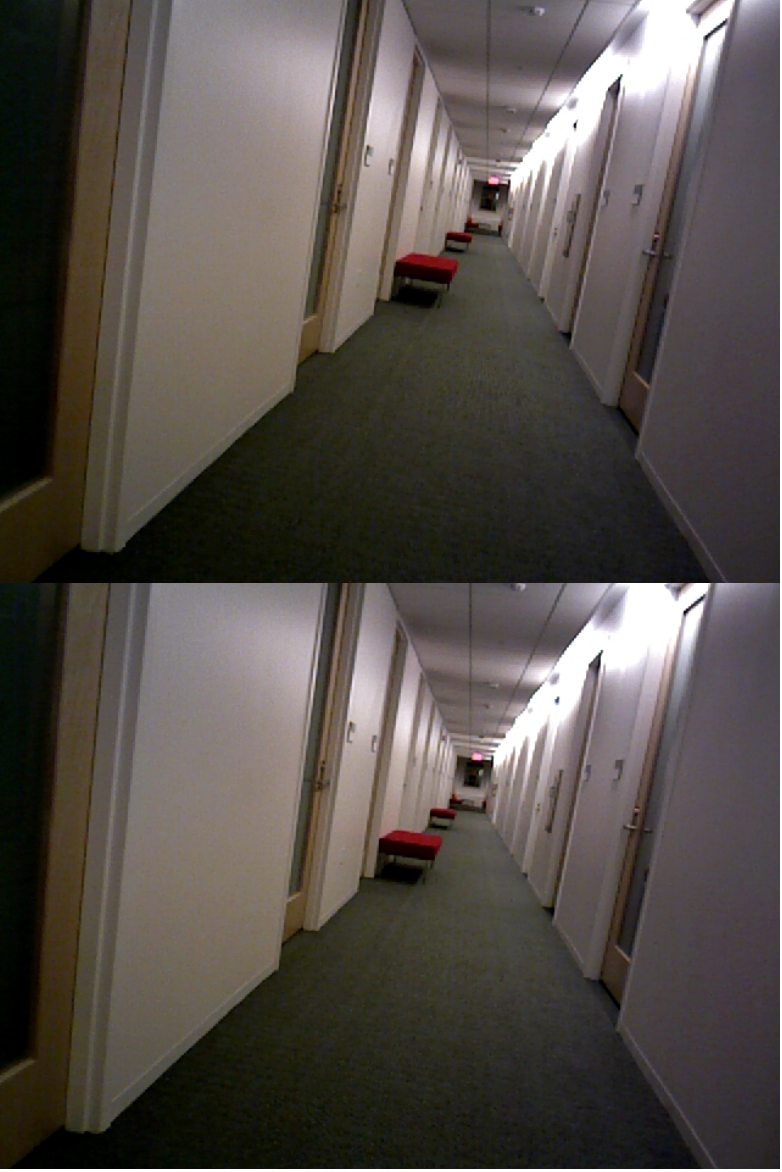}}
		\centerline{\includegraphics[width=2cm, height=3.65cm]{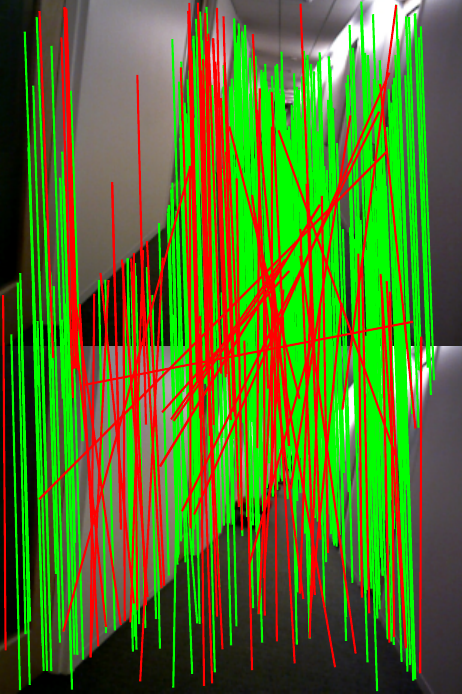}}
		\centerline{\includegraphics[width=2cm, height=3.65cm]{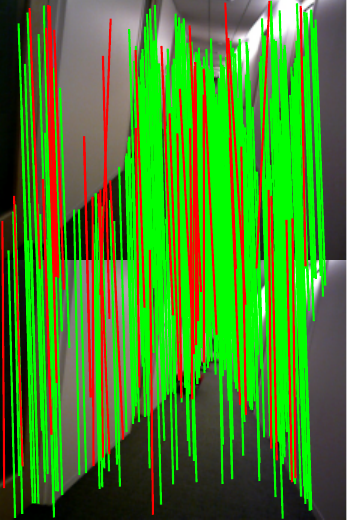}}
	\end{minipage}
		\hfill
	\caption{Partial typical visualization results on YFCC100M and SUN3D datasets with SIFT.     From top to bottom:  input image pairs, results of  CLNet and our MGNet. The green lines describe inliers, the red lines otherwise.
	}
	\label{vis}
\end{figure}

As shown in Table \ref{sift}, we present the quantitative results of the proposed MGNet  and main baselines on indoor and outdoor scenes with SIFT. Clearly, learning-based networks generally perform much better than traditional RANSAC. Besides, the proposed MGNet achieves the optimal value on all evaluation indicators. Our MGNet has increased $5.32$\% on mAP$5^{\circ}$ than the second best network (U-Match) in unknown outdoor scenes. Comparing to CL-Net, our MGNet gains performance increasements of $31.33$\%, $21.711$\%, $27.57$\% and $24.90$\% on the known outdoor, unknown outdoor, known indoor and unknown indoor scenes, respectively. And meanwhile, partial typical visualization results of CLNet and our MGNet in Figure \ref{vis} prove that the proposed MGNet can perform better under wide baseline,  large viewpoint changes, illumination changes and textureless region scenes. This is because our MGNet can effectively combine multiple graphs to capture and amplify discriminative features.

\subsubsection{Camera~Pose~Estimation~Results~with~SuperPoint.}
In addition, we choose a popular learning-based feature extractor, named SuperPoint \cite{detone2018superpoint}, to instead of SIFT to build putative correspondences. From Table \ref{superpoint}, we can find that our MGNet still achieves the best results in all situations. Comparing with Table \ref{sift} and Table \ref{superpoint}, there is a phenomenon that almost all methods (except for RANSAC and Point-Net++) perform better on the datasets preprocessed with SIFT than with SuperPoint.  Besides, for some performing poor methods (RANSAC and Point-Net++), more correct correspondences (SuperPoint) may have more advantageous. But, those networks, performing good enough, combine with SIFT better. That is explained in \cite{Dai_2022_CVPR}, in which Dai et al. prove that although SuperPoint obtains more correct correspondences than SIFT, but its average logit value is much lower.

\begin{table}[]
	\footnotesize
		\begin{tabular}
		{ccccc}
		\hline
		\multirow{2}*{Method}&\multicolumn{2}{c}{Outdoor(\%)} &\multicolumn{2}{c}{Indoor(\%)}  \\
		\cline{2-5}
		&\multicolumn{1}{c}{Known}&\multicolumn{1}{c}{Unknown}&\multicolumn{1}{c}{Known}&\multicolumn{1}{c}{Unknown} \\
		\cline{2-5}		
		\hline
		RANSAC  &   12.85   &  17.47&   14.93   &  12.15 \\	
		Point-Net++ &  11.87  &  17.95&11.40& 9.38\\
		DFE & 18.79 &  29.13&13.35& 12.04\\
		CNe  &12.18&24.25&12.63& 10.68\\			
		OA-Net++& 29.52 & 35.27 &20.01& 15.62\\
		ACNe  &26.72 &32.98& 18.35&13.82\\
		  CL-Net &29.35&38.99&15.89&14.03\\  
		  MS$^{2}$DG-Net  & 30.40  & 37.38&20.28 & 16.08    \\
		  U-Match &35.12&45.72&22.73&18.87	\\	
		  \cline{1-5}
		MGNet&  \textbf{41.53}&  \textbf{49.37}  &   \textbf{24.58}  &  \textbf{20.65}    \\			
			\hline		
	\end{tabular}
		\centering
	\caption{Evaluation on outdoor and indoor datasets with SuperPoint for camera pose estimation. mAP$5^{\circ}(\%)$ is reported.  }
	\label{superpoint}
\end{table}

\subsubsection{Generalization~Ability~Test.}
To evaluate the generalization ability of networks, we compare our MGNet and part of main baselines in different datasets with different feature extractors. Clearly, we introduce PhotoTourism  \cite{jin2021image} and ORB \cite{rublee2011orb} in the work, in which the former is a challenging photo-tourism dataset and the later is a fast yet accurate detector-descriptor method to be used as a preprocessing technique. We train all models on YFCC100M with SIFT and directly test them on different datasets with different extractors. As summarized in Table \ref{gen}, MGNet performs best in all setting, because it can effectively combine multiple different types of graphs to extract potential relationships among sparse correspondences. This can prove that MGNet has strong  robustness and generalization abilities.
\begin{table}[]
	\footnotesize
		\begin{tabular}
		{ccccc}
		\hline
		&\multicolumn{2}{c}{YFCC100M(\%)}  &\multicolumn{2}{c}{PhotoTourism(\%)} \\
		\cline{2-5}
		&\multicolumn{1}{c}{ORB}&\multicolumn{1}{c}{SP}&\multicolumn{1}{c}{SIFT}&\multicolumn{1}{c}{SP}  \\
		\cline{2-5}		
		\hline
		CNe  &7.40&14.78&20.17&5.89\\			
		OA-Net++ & 12.05 &19.40 &40.39&8.99 \\
		  CL-Net  &14.75&21.00&45.54&9.41\\  
		  MS$^{2}$DG-Net  & 11.38  &21.05& 45.53 &12.91    \\
		  U-Match &16.70&28.38&54.43 &11.48\\	
		  \cline{1-5}
		MGNet &  \textbf{20.00}&  \textbf{32.88}  & \textbf{57.64}  &  \textbf{20.41}     \\
		\hline		
	\end{tabular}
		\centering
	\caption{Generalization ability test on YFCC100M and PhotoTourism with different feature extractors, including ORB, SuperPoint (SP), and SIFT. mAP$5^{\circ}(\%)$ is reported.}
	\label{gen}
\end{table}
\subsection{Homography~Estimation}
The purpose of homography estimation is to find a linear image-to-image mapping in the homogeneous space, which is the basis for many subsequent computer vision tasks. We compare the proposed MGNet and part of main baselines on HPatches benchmark \cite{balntas2017hpatches} with Direct Linear Transform (DLT).  HPatches benchmark has $696$ images and $116$ scenes, each of which is composed of $1$ reference image and $5$ query images. That is, there are $580$ image pairs in HPatches benchmark, in which some are collected in viewpoint changes and others have different illumination. In our work, each image pair is detected up to 4000 keypoints with SIFT followed by a NN matcher. Following \cite{detone2018superpoint}, we choose homography error to evaluate them and present results that their average error is below $3/5/10$ pixels. Table \ref{hpatches} shows that MGNet performs best at all thresholds, especially obtains an absolute $3.18$  percentage point increase at the lowest threshold.

\subsection{Visual~Localization}
Visual localization is intended to estimate the 6-degree of freedom (DOF) camera pose of a given image relative to its 3D scene model, which is a fundamental problem in many computer vision and robotic tasks. Specifically, we integrate our MGNet and other comparative networks into the official HLoc \cite{sarlin2019coarse}. Aachen Day-Night benchmark \cite{sattler2018benchmarking} is chosen as a tested dataset, where $922$ query images ($824$ daytime and $98$ nighttime)  are captured by mobile phones and $4328$ reference ones are from a European ancient town. We extract up to $4096$ feature points with SIFT on each image, followed by a NN matcher. After that, a SfM model is triangulated from day-time images with known poses, and registers night-time query images with $2$D-$2$D matches obtained from correspondence learning networks and COLMAP \cite{schonberger2016structure}. Following HLoc \cite{sarlin2019coarse}, the percentage of correctly localized queries at specific distances and orientation thresholds is regarded as the evaluation matrix. Results in Table \ref{daynight} shows that our MGNet performs best in all situations and demonstrates that MGNet is suitable for visual localization.

\begin{table}[]
	\footnotesize
		\begin{tabular}
		{cccc}
		\hline
		\multirow{2}*{Method} &\multicolumn{3}{c}{HPathces(\%)}  \\
		\cline{2-4}
		&\multicolumn{1}{c}{ACC.\@3px}&\multicolumn{1}{c}{ACC.\@5px}&\multicolumn{1}{c}{ACC.\@10px} \\		
		\cline{2-4}		
		\hline
		CNe &38.97& 51.55 &65.34\\			
		OA-Net++ &39.83& 52.76& 62.93 \\
		  CL-Net  &43.10& 55.69& 68.10\\  
		  MS$^{2}$DG-Net &41.21& 50.17& 62.59 \\
		  U-Match &48.90&59.41&70.83\\
		  \cline{1-4}
		MGNet   &  \textbf{52.08}  & \textbf{61.53}  &  \textbf{71.23}      \\
		\hline		
	\end{tabular}
		\centering
	\caption{Evaluation homography estimation on HPatches.  }
	\label{hpatches}
\end{table}	

\begin{table}[]
	\footnotesize
		\begin{tabular}
		{ccc}
		\hline
		\centering		
		\multirow{2}*{Method} &\multicolumn{1}{c}{Day} &\multicolumn{1}{c}{Night} \\
		\cline{2-3}
		&\multicolumn{2}{c}{(0.25m, $2^{\circ}$)$/$(0.5m, $5^{\circ}$)$/$(1.0m, $10^{\circ}$)} \\		
		\cline{2-3}		
		\hline
		CNe&81.3/91.4/95.9&68.4/78.6/87.8\\			
		OA-Net++ & 82.3/91.9/96.5& 71.4/79.6/90.8\\
		  CL-Net  &83.3/92.4/ \textbf{97.0} &71.4/80.6/\textbf{93.9}\\  
		  MS$^{2}$DG-Net &82.8/92.1/96.8&70.4/82.7/\textbf{93.9}  \\
		  U-Match&\textbf{85.3}/92.6/96.8& 72.4/82.7/90.8	\\	
		  \cline{1-3}
		MGNet   &  \textbf{85.3/92.7/97.0}  & \textbf{ 72.6/82.9/93.9}        \\		
		\hline		
	\end{tabular}
	\centering
	\caption{Evaluation visual localization on Aachen Day-Night.}	
	\label{daynight}
\end{table}

\subsection{Ablation~Studies}
In this section, ablation studies about implicit graphs, explicit graphs,  global graphs, the relationship among them, the~pruning~operation and the~verification~framework on the unknown outdoor scene with SIFT are reported.
\subsubsection{How~to~choose~$m$?}
\begin{figure}
	\centering
	\includegraphics[width=5.6cm]{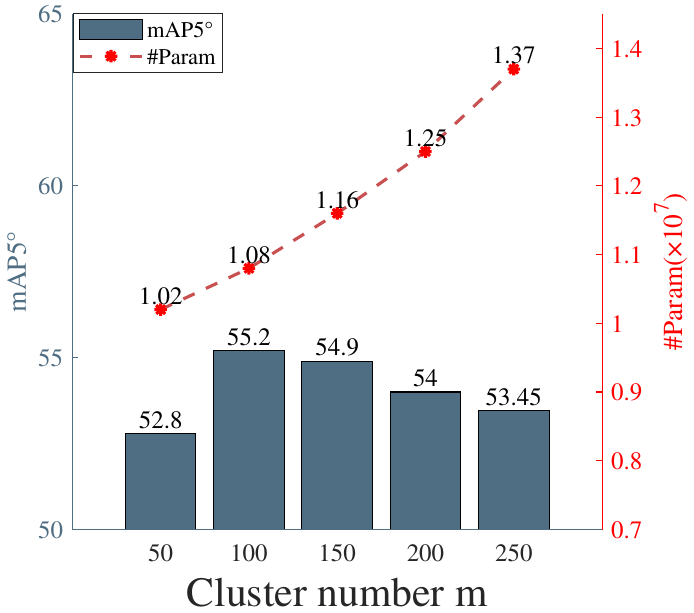}\\
	\caption{Relationship between mAP$5^{\circ}(\%)$  and network parameter number with different cluster number $m$. }\label{clusterm}
\end{figure}

Cluster number $m$, determining the coarsening degree of implicit graphs,  is pretty important. As shown in Figure \ref{clusterm}, in which coordinate axes on the left and right represent mAP$5^{\circ}$ under unknown YFCC100M scenes and the parameter number, respectively. And we can find that with cluster number $m$ increasing and after $m$ = $100$, the network performance gradually decreases while the parameter quantity still rises. Hence, we choose cluster number $m$ = $100$ to complete  subsequent experiments.

\subsubsection{How~to~choose~$k$?} 
\begin{figure}
	\centering
	\includegraphics[width=5.6cm]{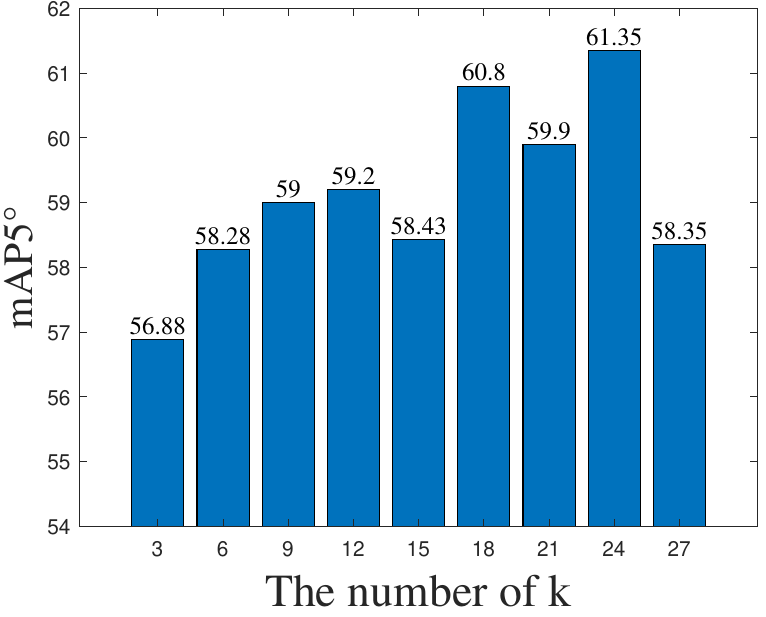}\\
	\caption{Parametric analysis of $k$ in the explicit local graph. }\label{numberk}
\end{figure}

The neighbor number  $k$, determining how much information in each explicit local graph,  is of vital importance. 
As shown in Figure \ref{numberk}, as  $k$ increases from $3$ to $27$, the network performance first increases and then decreases. When $k$ = $24$ is the turning point and also the best time for model performance, so we choose $k$ = $24$ to construct explicit local graphs.
\begin{table}[]
	\footnotesize	
		\begin{tabular}
		{c|ccc}
		\hline
		&\multicolumn{1}{c}{mAP$5^{\circ}$}&\multicolumn{1}{c}{mAP$20^{\circ}$}&\multicolumn{1}{c}{Size(MB)} \\		
		\cline{2-4}		
		\hline
		 Annular Conv&61.00& 81.13 &1.54\\			
		Avg-pooling\&MLPs&58.35& 79.62& \textbf{1.18} \\
		Max-pooling\&MLPs &  \textbf{61.35}  & \textbf{82.78}  &  \textbf{1.18}      \\		
		\hline		
	\end{tabular}
	\centering
	\caption{Quantitative comparisons of different aggregation methods in explicit local graphs.   }
	\label{agginformation}
\end{table}

\subsubsection{How~to~aggregate~information~in~explicit~graphs?}
We choose three ways (average pooling, maximum pooling and annular convolution \cite{zhao2021progressive})  to  fuse information.  And Table \ref{agginformation} denotes that maximum pooling method performs best on mAP$5^{\circ}(\%)$ and mAP$20^{\circ}(\%)$, which is simple yet effective. It is worth mentioning that average pooling and maximum pooling have the same number of parameters, but its results are worse. Annular convolution \cite{zhao2021progressive} not only has more parameters than maximum pooling, but also performs  poorer. Hence, we select  maximum pooling method. 	
\subsubsection{How~to~effectively~construct~global~graph~edge?} 
We choose four ways to construct~global~graph~edge. The first one is the plain GCN \cite{zhao2021progressive}, which is used in recent works. Combining with  Laplacian matrix knowledge, we propose the remaining three ways. Specifically, the second one is based on the Soft Adjacent Matrix, named GAA, and next one is to add a self-loop (its own information) on the second, called GAIA. The last one is GSDA, which is based on Soft Degree Matrix. As shown in Table \ref{agginformationg}, GSDA performs best and we choose it. 

\begin{table}[]
	\footnotesize
		\begin{tabular}
		{c|cccc}
		\hline
		&plain GCN &GAA&GAIA&GSDA\\	
		\cline{1-5}		
		mAP$5^{\circ}$&63.08&63.60&	64.00  &\textbf{64.63} \\		
		mAP$20^{\circ}$&82.72 &83.05&83.43  &\textbf{83.76} \\		 			
		\hline		
	\end{tabular}
		\centering
	\caption{Evaluate the plain GCN \cite{zhao2021progressive}, GAA, GAIA and GSDA to construct global edge.  }
	\label{agginformationg}
\end{table}

\subsubsection{Is~additional~local~information~helpful?}
As shown in Table \ref{eil}, we can find that only using the global probability to verify is enough, which not only minimizes the parameter number but also performs best. Besides, we also find the more local information (implicit or explicit) is added, the worse the network performs. It is probably because local information has already been integrated into the global, and reusing it can cause overfitting to degrade performance. 

\begin{table}[]
	\footnotesize	
		\begin{tabular}
		{c|ccc}
		\hline
		&\multicolumn{1}{c}{mAP$5^{\circ}$}&\multicolumn{1}{c}{mAP$20^{\circ}$}&\multicolumn{1}{c}{Size(MB)} \\		
		\cline{2-4}		
		\hline
		+$P_{li}$&61.83&82.04  &1.32\\
		+$P_{le}$&63.15& 83.09 &1.32\\			
		+$P_{li}$+$P_{le}$&61.23& 82.52& 1.42\\
		MGNet &  \textbf{64.63}  & \textbf{83.76}  &  \textbf{1.31}      \\			
		\hline		
	\end{tabular}
	\centering
	\caption{Compare effect of adding different local information. MGNet represents only utilize the global probability to verify. "+" represents add other information on MGNet. $P_{li}$ and $P_{le}$ represents implicit and explicit local probabilities.}
	\label{eil}
\end{table}

\begin{table}[]
\footnotesize
		\begin{tabular}
		{c|ccc}
		\hline
		&\multicolumn{1}{c}{mAP$5^{\circ}$}&\multicolumn{1}{c}{mAP$20^{\circ}$}&\multicolumn{1}{c}{Size(MB)} \\		
		\cline{2-4}		
		\hline	
		$wo$ verification &61.17& 81.64& 1.60\\
		 $w$ verification, $pr$=0.75 &50.80& 73.31&  \textbf{1.31} \\
		 $w$ verification, $pr$=0.5 &61.68& 81.69&  \textbf{1.31} \\
		$w$ verification, $pr$=0.25 &63.05&82.34&  \textbf{1.31} \\
		MGNet  &  \textbf{64.63}  & \textbf{83.76} &  \textbf{1.31}      \\			
		\hline		
	\end{tabular}
		\centering
	\caption{Parameter analysis of the pruning operation.  $pr$ presents the pruning ratio. $w$/$wo$ represent with/without. }	
	\label{pruning}
\end{table}\subsubsection{Is~the~pruning~operation~helpful?}
As summarized in Table \ref{pruning}, the model performance deteriorates with the pruning ratio increasing. Interestingly, when the pruning ratio is $0.25$, it performs worse than without  the verification framework. That is probably because if a model is not complex enough (without a large number of parameters), the pruning operation will reduce  data abundance (contrary to data augmentation), so that the model performance reduces.

\begin{table}[]
\footnotesize
	\footnotesize
		\begin{tabular}
		{c|ccc}
		\hline
		&\multicolumn{1}{c}{mAP$5^{\circ}$}&\multicolumn{1}{c}{mAP$20^{\circ}$}&\multicolumn{1}{c}{Size(MB)} \\		
		\cline{2-4}		
		\hline	
		 local explicit graph first&59.95& 81.92& \textbf{1.18}\\
		 local implicit graph first &  \textbf{61.35}  & \textbf{82.71}  & \textbf{1.18}   \\			
		\hline		
	\end{tabular}
		\centering
	\caption{Analysis about the order of building local graphs.  }	
	\label{relationship}
\end{table}
\subsubsection{Relationship~among~them.}
Comparing with Table \ref{relationship} and the third, fourth and fifth lines in Table \ref{final}, we find that building implicit graphs first is much better than building explicit graphs first, building implicit graphs twice and building explicit graphs twice. Observing the sixth, seventh and last lines in Table \ref{final}, we find that the combination we have been proposed (MGNet) performs much better than others, which can effectively combine local and global information.

\begin{table}[]
	\footnotesize
	\begin{tabular}
		{cccc|cc}
		\hline
		\multirow{1}*{Ver}&\multirow{1}*{IG}&\multirow{1}*{EG}&
		\multirow{1}*{GG}& \multicolumn{1}{c}{mAP$5^{\circ}$}&\multicolumn{1}{c}{mAP$20^{\circ}$} \\
		\cline{5-6}
		\hline
		$\checkmark$&& & &44.9&70.81    \\
		$\checkmark$&$\checkmark$& & &55.2 &77.24   \\
		$\checkmark$&$\checkmark$&$\checkmark$&&61.35& 82.71   \\
		$\checkmark$&$\checkmark$$\checkmark$&&&51.18& 75.06\\
		$\checkmark$&&$\checkmark$$\checkmark$&&59.78& 81.26\\
		$\checkmark$&$\checkmark$$\checkmark$$\checkmark$&&&51.43&75.47\\
		$\checkmark$&&$\checkmark$$\checkmark$$\checkmark$&&63.95&83.29\\
		&$\checkmark$&$\checkmark$&$\checkmark$&61.17& 81.64\\
		$\checkmark$&$\checkmark$&$\checkmark$&$\checkmark$& \textbf{64.63}  & \textbf{83.76} \\
		\hline
	\end{tabular}
	\centering
	\caption{Ablation studies about network compositions. Ver, IG, EG and GG represent the verification framework, implicit local graph, explicit local graph and global graph.}
	\label{final}
\end{table}
\subsubsection{Is~the~verification~framework~useful?} 
From the eighth and last lines in Table \ref{final}, we find the verification framework is very useful and can improve $3.46$ mAP$5^{\circ}(\%)$ on the model without it. This is probably  because using the verification framework can more fully explore potential relationships among sparse correspondences.

\section{Conclusion}
This work proposes MGNet for learning correspondences. There are two main improvements: 1) We construct local graphs from implicit and explicit aspects at the same time, and explore their potential relationship. 2) Motivated by Laplacian matrix, Graph~Soft~Degree~Attention (GSDA) is proposed to capture and amplify discriminative features based on the whole sparse correspondence information in the global graph. Experiments on different tasks and datasets prove that MGNet has a great performance improvement compared to other state-of-the-art methods. 

\section{Acknowledgments}
This work was supported by National Research Foundation, Singapore under its AI Singapore Programme (AISG Award No: AISG2-RP-2021-022), the National Natural Science Foundation of China (62172226), and the 2021 Jiangsu Shuangchuang (Mass Innovation and Entrepreneurship) Talent Program (JSSCBS20210200).

\bibliography{aaai24}

\end{document}